\pdfoutput=1
\documentclass[runningheads]{llncs}
\usepackage{graphicx}

\usepackage{hyperref}

\usepackage{cite}
\usepackage{amsmath}    
\usepackage{enumerate}  
\usepackage{rotating}
\usepackage{comment}

\usepackage{color,soul}
\usepackage{float}
\usepackage[table,xcdraw]{xcolor}
\usepackage{multirow}
\usepackage{hhline}
\usepackage{arydshln}

\usepackage{xspace}
\newcommand*{\eg}{\emph{e.g.}\@\xspace}
\newcommand*{\ie}{\emph{i.e.}\@\xspace}
\newcommand*{\proposed}{\textbf{Sli2Vol}\@\xspace}

\begin{document}
\title{Sli2Vol: Annotate a 3D Volume from a Single Slice with Self-Supervised Learning}
\titlerunning{Sli2Vol: Annotate a 3D Volume from a Single Slice}

\author{Pak-Hei Yeung\inst{1} \and
Ana I.L. Namburete\inst{1}$^{+}$\and
Weidi Xie\inst{1,2}$^{+}$}
%
\authorrunning{PH. Yeung et al.}
%
\institute{Department of Engineering Science, Institute of Biomedical Engineering, University of Oxford, Oxford, United Kingdom\\
\email{pak.yeung@pmb.ox.ac.uk, ana.namburete@eng.ox.ac.uk}
 \and
Visual Geometry Group, Department of Engineering Science, University of Oxford, Oxford, United Kingdom\\
\email{weidi@robots.ox.ac.uk}}

\maketitle              

\begin{center}
\url{https://pakheiyeung.github.io/Sli2Vol_wp/}
\end{center}

%
\begin{abstract}
The objective of this work is to segment any \emph{arbitrary} structures of interest (SOI) in 3D volumes by only annotating a \emph{single} slice,
(\ie~semi-automatic 3D segmentation).
We show that high accuracy can be achieved by simply propagating the 2D slice segmentation 
with an affinity matrix between consecutive slices, which can be learnt in a self-supervised manner, 
namely slice reconstruction.
Specifically, we compare our proposed framework, termed as \proposed,
with supervised approaches and two other unsupervised/ self-supervised slice registration approaches,
on 8 public datasets~(both CT and MRI scans), spanning 9 different SOIs.
Without any parameter-tuning,
the same model achieves superior performance with Dice scores (0-100 scale) of over $80$ for most of the benchmarks, including the ones that are unseen during training.
Our results show \emph{generalizability} of the proposed approach across data from different machines and with different SOIs: 
a major use case of semi-automatic segmentation methods where fully supervised approaches would normally struggle.





\end{abstract}
%
%
%
\section{Introduction}

\begin{figure*}
\centering
\includegraphics[width=\textwidth]{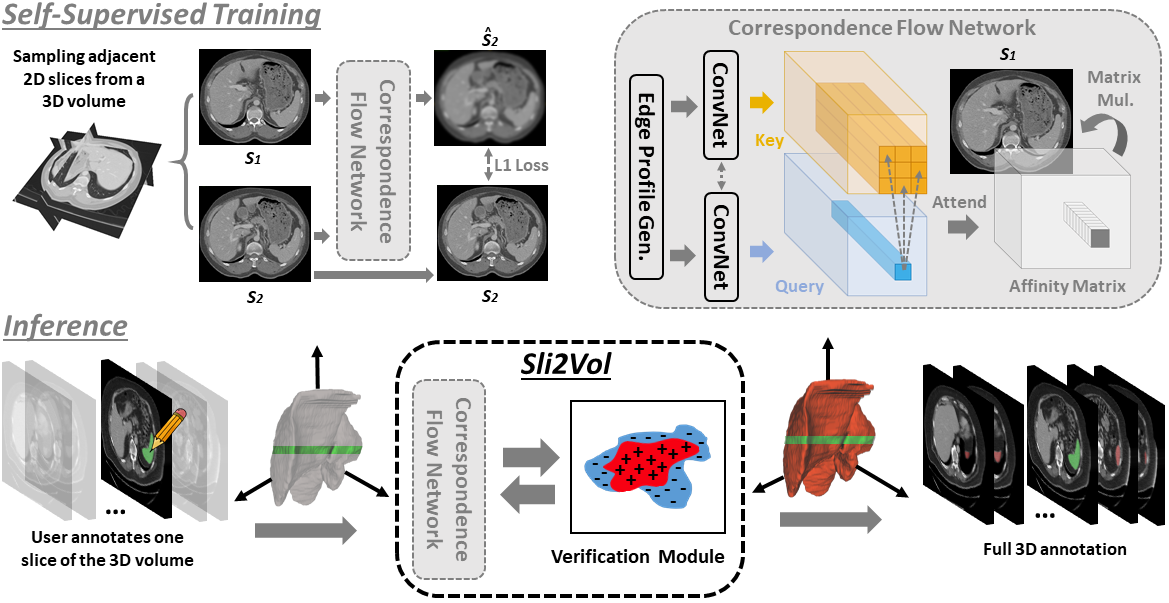}
\caption{Pipeline of our proposed framework. 
During \emph{self-supervised training}, pair of adjacent slices sampled from 3D volumes are used to train a correspondence flow network. 
Provided with the 2D mask of a single slice of a volume, the trained network with the verification module can be used to propagate the initial annotation to the whole volume during \emph{inference}.
} 
\label{pipeline}
\end{figure*}

Image segmentation is arguably one of the most important tasks in medical image analysis, 
as it identifies the structure of interest (SOI) with arbitrary shape~(\ie pixel level predictions), 
encompassing rich information, such as the position and size.
In recent years, the development and application of different deep convolutional neural networks (ConvNet), for example U-Net~\cite{Ronneberger15}, have significantly boosted the accuracy of computer-aided medical image segmentation.

Training fully automatic segmentation models comes with several limitations:
\emph{firstly}, annotations for the training volumes are usually a costly process to acquire;
\emph{secondly}, once domain shift appears, (\ie from differences in scanner, acquisition protocol or the SOI varies during inference), 
the model may suffer a catastrophic drop in performance, 
requiring new annotations and additional fine-tuning.
These factors have limited the use of the automatic segmentation approaches to
applications with inter-vendor and inter-operator variance.
 

As an alternative, semi-automatic approaches are able to operate interactively with the end users: 
this is the scenario considered in this paper.
Specifically, 
the goal is to segment any \emph{arbitrary} SOIs in 3D volumes by only annotating a \emph{single} slice within the volume, 
which may facilitate more flexible analysis of \emph{arbitrary} SOIs with the desired generalizability (\eg inter-scanner variability),
and significantly reduce the annotating cost for fully supervised learning.

Similar tools have been developed with level set or random forest methods, 
which show excellent performance as reported in~\cite{zheng2018unified, foruzan2016improved, li2013likelihood, dawant2007semi, wang2015slic}.
However, implementation of specific regularization and heavy parameter-tuning are usually required for different SOIs, 
limiting its use in practice.
On the other hand, related work in medical image registration explores the use of pixelwise correspondence from optical flow~\cite{hermann2013high, keeling2005medical} 
or unsupervised approaches~\cite{balakrishnan2019voxelmorph, mocanu2021flowreg, heinrich2013mrf}, 
which in principle could be harnessed for the propagation of a 2D mask between slices within a volume. 
However, they are prone to error drift, \ie error accumulation, introduced by inter-slice propagation of registration errors.
In this work, we aim to overcome these limitations.


Here, we focus on the task of propagating the 2D slice segmentation through the entire 3D volume by matching correspondences between consecutive slices. 
Our work makes the following contributions:
\emph{firstly}, we explore mask propagation approaches based on unsupervised/self-supervised registration of slices, 
namely,~na\"ive optical flow~\cite{farneback2003two} and VoxelMorph~\cite{balakrishnan2019voxelmorph}, 
and our proposed self-supervised approach, called \proposed, 
which is based on learning to match slices' correspondences~\cite{Lai19, lai2020mast} 
and using a newly proposed edge profile for information bottleneck.
\proposed is able to propagate the mask at a speed of 2.27 slices per second in inference.
\emph{Secondly}, to alleviate the error accumulation in mask propagation,
we propose and exploit a simple verification module for refining the mask during inference time.
\emph{Thirdly}, we benchmark \proposed on 8 public CT and MRI datasets~\cite{van07, CHAOSdata2019, soler20103d, simpson2019large}, spanning 9 anatomical structures.
Without any parameter-tuning,
a \emph{single} \proposed model achieves Dice scores (0-100 scale) above $80$ for most of the benchmarks,
which outperforms other supervised and unsupervised approaches for all datasets in cross-domain evaluation. 
To the best of our knowledge, 
this is the first study to undertake cross-domain evaluation on such large-scale and diverse benchmarks for semi-automatic segmentation approaches,
which shifts the focus to \emph{generalizability} across different devices, clinical sites and anatomical SOIs.

\label{sec:introduction}


\section{Methods}
\label{sec:method}


In Section~\ref{sec:setup}, 
we first formulate the problem setting in this paper, 
namely~semi-automatic segmentation for 3D volume with \emph{single} slice annotation.
Next, we introduce the training stage of our proposed approach, \proposed, in Section~\ref{sec:learning}
and our proposed edge profile generator in Section~\ref{sec:profile}.
This is followed by 
the computations for inference (\ref{sec:inference}), including our proposed verification module (\ref{sec:verification}).

\subsection{Problem Setup}
\label{sec:setup}
In general, given a 3D volume, denoted by $\mathbf{V} \in \mathcal{R}^{H \times W \times D}$, 
where $H$, $W$ and $D$ are the height, width and depth of the volume, respectively,
our goal is to segment the SOI in the volume based on a user-provided 2D segmentation mask for the \emph{single} slice,
\ie~$\mathbf{M}_i \in \mathcal{R}^{H \times W \times 1}$ with $1's$ indicating the SOI, and $0's$ as background.
The outputs will be a set of masks for an individual slice, \ie~$\{{\mathbf{M}_1}, {\mathbf{M}_2},...,{\mathbf{M}_D}\}$.

Inspired by~\cite{Lai19, lai2020mast},
we formulate this problem as learning feature representations that establish robust pixelwise correspondences between adjacent slices in a 3D volume, which results in a set of affinity matrices, 
$\mathbf{A}_{i \rightarrow i+1}$,
for propagating the 2D mask between consecutive slices by \emph{weighting and copying}.
Model training follows a self-supervised learning scheme,
where raw data is used, and only one slice annotation is required during inference time.

\subsection{Self-Supervised Training of Sli2Vol}
\label{sec:learning}
In this section, 
we detail the self-supervised approach for learning the dense correspondences.
Conceptually, 
the idea is to task a deep network for slice reconstruction by \emph{weighting and copying} pixels from its neighboring slice.
The affinity matrices used for weighting are acquired as a by-product, 
and can be directly used for mask propagation during inference.

During training, 
a pair of adjacent slices, $\{{\mathbf{S}_1}, {\mathbf{S}_2}\}, \mathbf{S}_i \in \mathcal{R}^{H \times W \times 1}$, 
are sampled from a training volume, 
and then fed to a ConvNet, parametrized by $\psi(\cdot,\theta)$~(as shown in the upper part of Fig.~\ref{pipeline}):
\begin{equation}
[\mathbf{k}_1,\ \mathbf{q}_2] = [\psi(g(\mathbf{S}_1) ;\ \theta),\ \psi(g(\mathbf{S}_2) ;\ \theta)]
\end{equation}
where $g(\cdot)$ denotes an \emph{edge profile generator}~(details in Section~\ref{sec:profile}) 
and $\mathbf{k}_1, \mathbf{q}_2 \in \mathcal{R}^{H \times W \times c}$ 
refer to the feature representation~($c$ channels) computed from corresponding slices, 
termed as \emph{key} and \emph{query} respectively (Fig.~\ref{pipeline}).
The difference in notation (\ie $\mathbf{q}$ and $\mathbf{k}$) is just for emphasizing their functional difference.


Reshaping $\mathbf{k}_1$ and $\mathbf{q}_2$ to $\mathcal{R}^{HW \times c}$, 
an affinity matrix, $\mathbf{A}_{1 \rightarrow 2} \in \mathcal{R}^{HW \times \delta}$, 
is computed to represent the feature similarity between the two slices (Fig.~\ref{pipeline}):
\begin{equation}
\mathbf{A}_{1 \rightarrow 2}(u, v) = \frac{exp\langle \mathbf{q}_2(u,:), \mathbf{k}_1(v,:) \rangle}{\sum_{\lambda \in \Omega}{exp\langle \mathbf{q}_2(u,:), \mathbf{k}_1(\lambda,:) \rangle}}
\end{equation}
where $\langle\cdot{,}\cdot\rangle$ is the dot product between two vectors 
and $\Omega$ is the window surrounding pixel $v$ (\ie in $\mathcal{R}^{H \times W}$ space) for computing local attention,
with $n(\Omega) = \delta$.
%
\subsubsection{Loss Function.}
During training, $\mathbf{A}_{1 \rightarrow 2}$ is used to \emph{weight and copy} pixels from $\mathbf{S}_1$ (\ie reshape to $\mathcal{R}^{HW \times 1}$) to reconstruct $\mathbf{S}_2$, denoted as $\hat{\mathbf{S}}_2$, by:
%
\begin{equation}
\label{equ:propagate}
\hat{\mathbf{S}}_2(u,1) = \sum_v^\Omega{\mathbf{A}_{1 \rightarrow 2}(u,v)\mathbf{S}_1(v,1).}
\end{equation}
\noindent 
We apply mean absolute error (MAE) between $\mathbf{S}_2$ and $\hat{\mathbf{S}}_2$ as the training loss.
%

\subsection{Edge Profile Generator}
\label{sec:profile}
Essentially, the basic assumption of the above-mentioned idea is that, 
to better reconstruct $\mathbf{S}_2$ via copying pixel from $\mathbf{S}_1$, 
the model must learn to establish reliable correspondences between the two slices. 
However, na\"ively training the model may actually incur trivial solutions, 
for example, the model can perfectly solve the reconstruction task by simply matching the \emph{pixel intensity} of $\mathbf{S}_1$ and $\mathbf{S}_2$. 

In Lai \emph{et al.}~\cite{Lai19, lai2020mast}, the authors show that 
input color channel (\ie \emph{RGB} or \emph{Lab}) dropout is an effective information bottleneck,
which breaks the correlation between the color channels and forces the model to learn more robust correspondences.
However, this is usually not feasible in medical images, as only single input channel is available in most of the modalities.

We propose to use a \emph{profile of edges} as an \emph{information bottleneck} to avoid trivial solution.
Specifically, for each pixel, we convert its intensity value to a normalized edge histogram,
by computing the derivatives along $d$ different directions at $s$ different scales, 
\emph{i.e.}~$g(\mathbf{S}_i) \in \mathcal{R}^{H \times W \times (d \times s)}$,
followed by a \emph{softmax} normalization through all the derivatives.
Intuitively, $g(\cdot)$ explicitly represents the edge distributions centered each pixel of the slice $\mathbf{S}_i$,
and force the model to pay more attentions to the edges during reconstruction.
Experimental results in Section~\ref{sec:result} verify the essence of this design in improving the model performance.


\subsection{Inference}
\label{sec:inference}
Given a volume, $\mathbf{V}$ and an initial mask at the $i$-th slice, $\mathbf{M}_i$, 
the affinity matrix, $\mathbf{A}_{i \rightarrow i+1}$, output from $\psi(\cdot,\theta)$ 
is used to propagate $\mathbf{M}_i$ iteratively to the whole $\mathbf{V}$.

In detail,
two consecutive slices, $\{{\mathbf{S}_i}, {\mathbf{S}_{i+1}}\}$, 
are sampled from the volume $\mathbf{V}$ and fed into $\psi(g(\cdot),\theta)$ to get $\mathbf{A}_{i \rightarrow i+1}$, 
which is then used to propagate $\mathbf{M}_i$, using Eq.~\ref{equ:propagate},
ending up with $\mathbf{\hat{M}}_{i+1}$.
This set of computations 
(Fig.~\ref{refine} 
in the Supplementary Materials in Section~\ref{sec:supp})
is then repeated for the next two consecutive slices, 
$\{{\mathbf{S}_{i+1}}, {\mathbf{S}_{i+2}}\}$, 
in either direction, 
until the whole volume is covered.

\subsection{Verification Module}
\label{sec:verification}
In practice, 
we find that directly using $\mathbf{\hat{M}}_{i+1}$ for further propagation will potentially accumulate the prediction error after each iteration.
To alleviate this drifting issue, and further boost the performance,
we propose a simple verification module to correct the mask after each iteration of mask propagation.

Specifically, two regions, 
namely positive~($\mathbf{P} \in \mathcal{R}^{H \times W}$) and negative~($\mathbf{N} \in \mathcal{R}^{H \times W}$) regions, are constructed.
$\mathbf{P}$ refers to the delineated SOI in $\mathbf{M}_i$,
and $\mathbf{N}$ is identified by subtracting $\mathbf{P}$ from its own morphologically dilated version.
Intuitively, the negative region denotes the thin and non-overlapping region surrounding $\mathbf{P}$ 
(Fig.~\ref{refine} 
in the Supplementary Materials in Section~\ref{sec:supp}).
We maintain the \emph{mean intensity value} within each region:
%
%
\begin{align*}
p = \frac{1}{|P_i|} \langle  P_i, S_i \rangle \text{\hspace{40pt}}
n = \frac{1}{|N_i|} \langle  N_i, S_i \rangle
\end{align*}
where $\langle \cdot, \cdot \rangle$ denotes Frobenius inner product,
$p$ and $n$ refer to the positive and negative query values respectively. 

During inference time, assuming $\mathbf{\hat{M}}_{i+1}$ is the predicted mask from the propagation,
each of the proposed foreground pixels $u$ in $\mathbf{S}_{i+1}$, 
is then compared to $p$ and $n$ 
and being re-classified according to its distance to the two values by:
%
\begin{equation}
  \mathbf{M}_{i+1}^u=\left\{
  \begin{array}{@{}ll@{}}
    1,\ \ \  & \text{if}\ \mathbf{\hat{M}}_{i+1}^u=1\ \text{and}\ \sqrt{(\mathbf{S}_{i+1}^u-p)^2}<\sqrt{(\mathbf{S}_{i+1}^u-n)^2}
 \\
    0,\ \ \  & \text{otherwise}
  \end{array}\right.
\end{equation} 
This set of computations 
is then repeated for the next round of propagation,
where $p$ and $n$ are updated using the corrected mask, $\mathbf{M}_{i+1}$, and $\mathbf{S}_{i+1}$.


%



\section{Experimental Setup}
We benchmark our framework, \proposed, on 8 different public datasets, spanning 9 different SOIs,
and compare with a variety of fully supervised and semi-automatic approaches, using standard Dice coefficient (in a 0-100 scale) as the evaluation metrics.
In Section~\ref{sec:data}, we introduce the datasets used in this paper.
In Section~\ref{sec:design}, we summarize the experiments conducted for this study.

\subsection{Dataset}
\label{sec:data}
Four training and eight testing datasets are involved.
For \textbf{chest and abdominal CT}, 
a \emph{single} model is trained on 3 unannotated dataset ($i.e.$ C4KC-KiTS~\cite{c4kc}, CT-LN~\cite{lymphnodes} 
and CT-Pancreas~\cite{pancreasct}) 
and tested on 7 other datasets ($i.e.$ Sliver07~\cite{van07}, CHAOS~\cite{CHAOSdata2019}, 
3Dircadb-01, 02~\cite{soler20103d}, 
and Decath-Spleen, Liver and Pancreas~\cite{simpson2019large}).


For \textbf{cardiac MRI}, models are trained on the 2D video dataset from Kaggle~\cite{kaggle}, 
and tested on a 3D volume dataset ($i.e.$ Decath-Heart~\cite{simpson2019large}),
which manifests large domain shift.
Further details of the datasets are provided in 
Table~\ref{table:data} 
in the Supplementary Materials in Section~\ref{sec:supp}.

\subsection{Baseline Comparison}
\label{sec:design}
\proposed and a set of baseline approaches are tested, 
with their implementation details summarized in 
Table~\ref{table:implementation} 
in the Supplementary Materials in Section~\ref{sec:supp}.

$First$, we experiment with two approaches trained on fully annotated 3D data.
\textbf{Fully Supervised (FS) - Same Domain} refers to the scenario where the training and testing data come from the $same$ benchmark dataset.
Results from both state-of-the-art methods~\cite{Isensee19, ahmad2019deep, kavur2020chaos, tran2020multiple}
and 3D UNets trained by us are reported.
On the other hand, 
\textbf{FS - Different Domain} aims to evaluate the generalizability of FS approaches 
when training and testing data come from \emph{different} domains.
Therefore, 
we train the 3D UNet (same architecture and settings as the \textbf{FS - Same Domain} for fair comparison)
on a source dataset, 
and test it on another benchmark of the same task.


\emph{Second}, 
we consider the case where only a single slice is annotated in each testing volume to train a 2D UNet (\textbf{FS - Single Slice}).
For example, 
in \emph{Sliver07},
the model trained on 20 slice annotations~(single slice from each volume),
is tested on the same set of 20 volumes.
This approach utilizes the same amount of manual annotation as \proposed, 
so as to investigate if a model trained on single slice annotations is sufficient to generalize to the whole volume.

\emph{Third}, approaches based on registration of consecutive slices, 
namely \textbf{Optical Flow}~\cite{farneback2003two, opencv_library}, \textbf{VoxelMorph2D (VM) - UNet} and \textbf{VM - ResNet18Stride1}, are tested. 
The two VMs utilize a UNet backbone as proposed originally in~\cite{balakrishnan2019voxelmorph} 
as well as the same backbone (\ie ResNet18Stride1) as \proposed, respectively.

For \proposed, \textbf{FS - Single Slice}, \textbf{Optical Flow} and \textbf{VM}, 
we randomly pick one of the $\pm 3$ slices around the slice with the largest groundtruth annotation as the initial mask.
This simulates the process of a user sliding through the whole volume and roughly identifying the slice with the largest SOI to annotate, which is achievable in reality.

\label{sec:exp}

\begin{sidewaystable}
\fontsize{7}{9}\selectfont
\begin{tabular}{|l|l|l|l|l|l|l|l|l|l|l|l|l|l|l|c|}
\hline
\multicolumn{1}{|c|}{\textit{\textbf{Modality}}}                                                                     & \textit{\textbf{\begin{tabular}[c]{@{}l@{}}MRI\end{tabular}}}   & \multicolumn{13}{l|}{\textit{\textbf{Abdominal and Chest CT}}}                                                                                                                                                                                                                                                                                                                                                                                                                                                                                                                                                                                                                                                                                                                                                                                                                                                                                            &                                                                          \\ \cline{1-15}
\multicolumn{1}{|c|}{\textit{\textbf{\begin{tabular}[c]{@{}c@{}}Training Dataset \\ (for row e to j)\end{tabular}}}} & \textit{\textbf{Kaggle}}                                                  & \multicolumn{13}{c|}{\textit{\textbf{C4KC-KiTS, CT-LN and CT-Pancreas}}}                                                                                                                                                                                                                                                                                                                                                                                                                                                                                                                                                                                                                                                                                                                                                                                                                                                                                  & \multicolumn{1}{l|}{}                                                    \\ \cline{1-15}
\multicolumn{1}{|c|}{\textit{\textbf{Testing Dataset}}}                                                              & \textit{\textbf{\begin{tabular}[c]{@{}l@{}}Decath\\ -Heart\end{tabular}}} & \textit{\textbf{Sliver07}}                                       & \textit{\textbf{CHAOS}}                                          & \textit{\textbf{\begin{tabular}[c]{@{}l@{}}Decath\\ -Liver\end{tabular}}} & \textit{\textbf{\begin{tabular}[c]{@{}l@{}}Decath\\ -Spleen\end{tabular}}} & \textit{\textbf{\begin{tabular}[c]{@{}l@{}}Decath-\\ Pancreas\end{tabular}}} & \multicolumn{8}{l|}{\textit{\textbf{3D-IRCADb-01 and 3D-IRCADb-02}}}                                                                                                                                                                                                                                                                                                                                                                                                                                                                                                        &                                                                          \\ \cline{1-15}
\multicolumn{1}{|c|}{\textit{\textbf{ROI}}}                                                                          & \textit{\textbf{\begin{tabular}[c]{@{}l@{}}Left\\ Atrium\end{tabular}}}   & \textit{\textbf{Liver}}                                          & \textit{\textbf{Liver}}                                          & \textit{\textbf{Liver}}                                                   & \textit{\textbf{Spleen}}                                                   & \textit{\textbf{Pancreas}}                                                   & \textit{\textbf{Heart}}                                           & \textit{\textbf{\begin{tabular}[c]{@{}l@{}}Gall-\\ bladder\end{tabular}}} & \textit{\textbf{Kidney}}                                         & \textit{\textbf{\begin{tabular}[c]{@{}l@{}}Surrenal-\\ gland\end{tabular}}} & \textit{\textbf{Liver}}                                          & \textit{\textbf{Lung}}                                            & \textit{\textbf{Pancreas}}                                       & \textit{\textbf{Spleen}}                                         & \textit{\textbf{\begin{tabular}[c]{@{}c@{}}Mean\\ Results\end{tabular}}} \\ \cline{1-15}
\multicolumn{1}{|c|}{\textit{\textbf{Number of Volumes}}}                                                            & \textit{\textbf{20}}                                                      & \textit{\textbf{20}}                                             & \textit{\textbf{20}}                                             & \textit{\textbf{131}}                                                     & \textit{\textbf{41}}                                                       & \textit{\textbf{281}}                                                        & \textit{\textbf{3}}                                               & \textit{\textbf{8}}                                                       & \textit{\textbf{17}}                                             & \textit{\textbf{11}}                                                        & \textit{\textbf{22}}                                             & \textit{\textbf{12}}                                              & \textit{\textbf{4}}                                              & \textit{\textbf{7}}                                              &                                                                          \\ \hline
\multicolumn{16}{|c|}{\cellcolor[HTML]{EFEFEF}Automatic (Trained with Fully Annotated Data)}                                                                                                                                                                                                                                                                                                                                                                                                                                                                                                                                                                                                                                                                                                                                                                                                                                                                                                                                                                                                                                                                                                                                            \\ \hline
\begin{tabular}[c]{@{}l@{}}(a) Fully Supervised-\\ same domain\end{tabular}                                          & 92.7\cite{Isensee19}                                                                                                                              & \begin{tabular}[c]{@{}l@{}}94.8\cite{ahmad2019deep}\\ (93.9)\end{tabular}                & \begin{tabular}[c]{@{}l@{}}97.8\cite{kavur2020chaos}\\ (92.8)\end{tabular}                & \begin{tabular}[c]{@{}l@{}}95.4\cite{Isensee19}\\ (91.0)\end{tabular}                     & 96.0\cite{Isensee19}                                                                       & 79.3\cite{Isensee19}                                                                         & \multicolumn{1}{c|}{-}                                            & \multicolumn{1}{c|}{-}                                                    & \multicolumn{1}{c|}{-}                                           & \multicolumn{1}{c|}{-}                                                      & 96.5\cite{tran2020multiple} & \multicolumn{1}{c|}{-}                                            & \multicolumn{1}{c|}{-}                                           & \multicolumn{1}{c|}{-}                                           & -                                                                        \\ \hline
\begin{tabular}[c]{@{}l@{}}(b) Fully Supervised-\\ different domain\end{tabular}                                     & \multicolumn{1}{c|}{-}                                                    & \begin{tabular}[c]{@{}l@{}}74.8\\ $\pm$13.2\end{tabular}         & \begin{tabular}[c]{@{}l@{}}76.5\\ $\pm$8.8\end{tabular}          & \begin{tabular}[c]{@{}l@{}}56.0\\ $\pm$23.6\end{tabular}                  & \multicolumn{1}{c|}{-}                                                     & \multicolumn{1}{c|}{-}                                                       & \multicolumn{1}{c|}{-}                                            & \multicolumn{1}{c|}{-}                                                    & \multicolumn{1}{c|}{-}                                           & \multicolumn{1}{c|}{-}                                                      & \multicolumn{1}{c|}{-}                                           & \multicolumn{1}{c|}{-}                                            & \multicolumn{1}{c|}{-}                                           & \multicolumn{1}{c|}{-}                                           & -                                                                        \\ \hline
\multicolumn{16}{|c|}{\cellcolor[HTML]{EFEFEF}Semi-automatic}                                                                                                                                                                                                                                                                                                                                                                                                                                                                                                                                                                                                                                                                                                                                                                                                                                                                                                                                                                                                                                                                                                                                                                           \\ \hline
\begin{tabular}[c]{@{}l@{}}(c) Fully Supervised-\\ single slice\end{tabular}                                         & \begin{tabular}[c]{@{}l@{}}62.5\\ $\pm$5.2\end{tabular}                   & \begin{tabular}[c]{@{}l@{}}86.9\\ $\pm$4.1\end{tabular}          & \begin{tabular}[c]{@{}l@{}}84.3\\ $\pm$4.1\end{tabular}          & \begin{tabular}[c]{@{}l@{}}85.0\\ $\pm$5.5\end{tabular}                   & \begin{tabular}[c]{@{}l@{}}74.4\\ $\pm$12.0\end{tabular}                   & \begin{tabular}[c]{@{}l@{}}49.9\\ $\pm$13.4\end{tabular}                     & \begin{tabular}[c]{@{}l@{}}25.6\\ $\pm$6.5\end{tabular}           & \begin{tabular}[c]{@{}l@{}}47.9\\ $\pm$15.5\end{tabular}                  & \begin{tabular}[c]{@{}l@{}}57.9\\ $\pm$21.1\end{tabular}         & \begin{tabular}[c]{@{}l@{}}30.8\\ $\pm$15.6\end{tabular}                    & \begin{tabular}[c]{@{}l@{}}80.3\\ $\pm$13.8\end{tabular}         & \begin{tabular}[c]{@{}l@{}}81.0\\ $\pm$10.8\end{tabular}          & \begin{tabular}[c]{@{}l@{}}20.4\\ $\pm$7.9\end{tabular}          & \begin{tabular}[c]{@{}l@{}}58.6\\ $\pm$4.7\end{tabular}          & 60.4                                                                     \\ \hline
(d) Optical Flow                                                                                                     & \begin{tabular}[c]{@{}l@{}}51.1\\ $\pm$7.4\end{tabular}                   & \begin{tabular}[c]{@{}l@{}}65.2\\ $\pm$8.8\end{tabular}          & \begin{tabular}[c]{@{}l@{}}72.0\\ $\pm$9.9\end{tabular}          & \begin{tabular}[c]{@{}l@{}}47.0\\ $\pm$15.9\end{tabular}                  & \begin{tabular}[c]{@{}l@{}}72.9\\ $\pm$14.5\end{tabular}                   & \begin{tabular}[c]{@{}l@{}}25.1\\ $\pm$8.2\end{tabular}                      & \begin{tabular}[c]{@{}l@{}}32.2\\ $\pm$11.6\end{tabular}          & \begin{tabular}[c]{@{}l@{}}24.6\\ $\pm$12.4\end{tabular}                  & \begin{tabular}[c]{@{}l@{}}73.6\\ $\pm$14.6\end{tabular}         & \begin{tabular}[c]{@{}l@{}}22.1\\ $\pm$12.9\end{tabular}                    & \begin{tabular}[c]{@{}l@{}}68.4\\ $\pm$9.4\end{tabular}          & \begin{tabular}[c]{@{}l@{}}33.6\\ $\pm$18.0\end{tabular}          & \begin{tabular}[c]{@{}l@{}}21.9\\ $\pm$12.6\end{tabular}         & \begin{tabular}[c]{@{}l@{}}70.8\\ $\pm$17.5\end{tabular}         & 48.6                                                                     \\ \hline
\begin{tabular}[c]{@{}l@{}}(e) VoxelMorph2D-\\ UNet\end{tabular}                                                     & \begin{tabular}[c]{@{}l@{}}42.9\\ $\pm$5.0\end{tabular}                   & \begin{tabular}[c]{@{}l@{}}57.2\\ $\pm$9.8\end{tabular}          & \begin{tabular}[c]{@{}l@{}}66.5\\ $\pm$10.5\end{tabular}         & \begin{tabular}[c]{@{}l@{}}38.5\\ $\pm$12.5\end{tabular}                  & \begin{tabular}[c]{@{}l@{}}61.5\\ $\pm$19.5\end{tabular}                   & \begin{tabular}[c]{@{}l@{}}21.4\\ $\pm$6.7\end{tabular}                      & \begin{tabular}[c]{@{}l@{}}20.3\\ $\pm$6.5\end{tabular}           & \begin{tabular}[c]{@{}l@{}}20.2\\ $\pm$12.2\end{tabular}                  & \begin{tabular}[c]{@{}l@{}}70.1\\ $\pm$18.6\end{tabular}         & \begin{tabular}[c]{@{}l@{}}41.1\\ $\pm$15.3\end{tabular}                    & \begin{tabular}[c]{@{}l@{}}60.5\\ $\pm$9.7\end{tabular}          & \begin{tabular}[c]{@{}l@{}}38.7\\ $\pm$21.2\end{tabular}          & \begin{tabular}[c]{@{}l@{}}28.3\\ $\pm$11.0\end{tabular}         & \begin{tabular}[c]{@{}l@{}}54.1\\ $\pm$12.4\end{tabular}         & 44.4                                                                     \\ \hline
\begin{tabular}[c]{@{}l@{}}(f) VoxelMorph2D-\\ ResNet18NoStride\end{tabular}                                         & \begin{tabular}[c]{@{}l@{}}45.7\\ $\pm$4.1\end{tabular}                   & \begin{tabular}[c]{@{}l@{}}61.2\\ $\pm$8.5\end{tabular}          & \begin{tabular}[c]{@{}l@{}}68.4\\ $\pm$9.8\end{tabular}          & \begin{tabular}[c]{@{}l@{}}42.2\\ $\pm$12.4\end{tabular}                  & \begin{tabular}[c]{@{}l@{}}58.3\\ $\pm$17.3\end{tabular}                   & \begin{tabular}[c]{@{}l@{}}23.5\\ $\pm$7.8\end{tabular}                      & \begin{tabular}[c]{@{}l@{}}22.1\\ $\pm$6.7\end{tabular}           & \begin{tabular}[c]{@{}l@{}}21.8\\ $\pm$13.1\end{tabular}                  & \begin{tabular}[c]{@{}l@{}}77.8\\ $\pm$18.4\end{tabular}         & \begin{tabular}[c]{@{}l@{}}48.4\\ $\pm$15.3\end{tabular}                    & \begin{tabular}[c]{@{}l@{}}60.6\\ $\pm$10.4\end{tabular}         & \begin{tabular}[c]{@{}l@{}}36.5\\ $\pm$20.0\end{tabular}          & \begin{tabular}[c]{@{}l@{}}32.3\\ $\pm$13.3\end{tabular}         & \begin{tabular}[c]{@{}l@{}}60.0\\ $\pm$12.1\end{tabular}         & 47.5                                                                     \\ \hline
\proposed                                                                                                     & \multicolumn{15}{l|}{\textit{\textbf{Ablation Studies}}}                                                                                                                                                                                                                                                                                                                                                                                                                                                                                                                                                                                                                                                                                                                                                                                                                                                                                                                                                                                                                                                         \\ \hdashline
\begin{tabular}[c]{@{}l@{}}(g) Correspondence\\ Flow Network\end{tabular}                                            & \begin{tabular}[c]{@{}l@{}}62.4\\ $\pm$9.2\end{tabular}                   & \begin{tabular}[c]{@{}l@{}}75.0\\ $\pm$6.5\end{tabular}          & \begin{tabular}[c]{@{}l@{}}78.9\\ $\pm$7.9\end{tabular}          & \begin{tabular}[c]{@{}l@{}}66.0\\ $\pm$13.1\end{tabular}                  & \begin{tabular}[c]{@{}l@{}}81.1\\ $\pm$13.9\end{tabular}                   & \begin{tabular}[c]{@{}l@{}}43.9\\ $\pm$12.9\end{tabular}                     & \begin{tabular}[c]{@{}l@{}}55.4\\ $\pm$24.3\end{tabular}          & \begin{tabular}[c]{@{}l@{}}62.4\\ $\pm$20.7\end{tabular}                  & \begin{tabular}[c]{@{}l@{}}86.0\\ $\pm$19.0\end{tabular}         & \begin{tabular}[c]{@{}l@{}}45.9\\ $\pm$18.6\end{tabular}                    & \begin{tabular}[c]{@{}l@{}}75.0\\ $\pm$8.6\end{tabular}          & \begin{tabular}[c]{@{}l@{}}45.2\\ $\pm$25.4\end{tabular}          & \begin{tabular}[c]{@{}l@{}}44.3\\ $\pm$17.2\end{tabular}         & \begin{tabular}[c]{@{}l@{}}81.8\\ $\pm$19.6\end{tabular}         & 64.5                                                                     \\ \hdashline
\begin{tabular}[c]{@{}l@{}}(h) Network + Edge \\ Profile\end{tabular}                                                & \begin{tabular}[c]{@{}l@{}}56.8\\ $\pm$8.4\end{tabular}                   & \begin{tabular}[c]{@{}l@{}}74.8\\ $\pm$7.4\end{tabular}          & \begin{tabular}[c]{@{}l@{}}77.8\\ $\pm$8.4\end{tabular}          & \begin{tabular}[c]{@{}l@{}}64.4\\ $\pm$14.1\end{tabular}                  & \begin{tabular}[c]{@{}l@{}}83.6\\ $\pm$13.2\end{tabular}                   & \begin{tabular}[c]{@{}l@{}}48.9\\ $\pm$11.2\end{tabular}                     & \begin{tabular}[c]{@{}l@{}}49.4\\ $\pm$12.3\end{tabular}          & \begin{tabular}[c]{@{}l@{}}68.5\\ $\pm$13.8\end{tabular}                  & \begin{tabular}[c]{@{}l@{}}86.8\\ $\pm$15.7\end{tabular}         & \textbf{\begin{tabular}[c]{@{}l@{}}58.3\\ $\pm$16.6\end{tabular}}           & \begin{tabular}[c]{@{}l@{}}73.9\\ $\pm$8.5\end{tabular}          & \begin{tabular}[c]{@{}l@{}}48.8\\ $\pm$26.4\end{tabular}          & \begin{tabular}[c]{@{}l@{}}53.9\\ $\pm$7.1\end{tabular}          & \begin{tabular}[c]{@{}l@{}}85.8\\ $\pm$13.0\end{tabular}         & 66.6                                                                     \\ \hdashline
\begin{tabular}[c]{@{}l@{}}(i) Network + Verif. \\ Module\end{tabular}                                               & \textbf{\begin{tabular}[c]{@{}l@{}}80.8\\ $\pm$5.0\end{tabular}}          & \begin{tabular}[c]{@{}l@{}}81.1\\ $\pm$5.0\end{tabular}          & \begin{tabular}[c]{@{}l@{}}83.4\\ $\pm$6.3\end{tabular}          & \begin{tabular}[c]{@{}l@{}}72.0\\ $\pm$8.9\end{tabular}                   & \begin{tabular}[c]{@{}l@{}}79.1\\ $\pm$17.3\end{tabular}                   & \begin{tabular}[c]{@{}l@{}}37.3\\ $\pm$13.6\end{tabular}                     & \begin{tabular}[c]{@{}l@{}}50.9\\ $\pm$11.6\end{tabular}          & \begin{tabular}[c]{@{}l@{}}70.7\\ $\pm$12.7\end{tabular}                  & \begin{tabular}[c]{@{}l@{}}83.3\\ $\pm$21.4\end{tabular}         & \begin{tabular}[c]{@{}l@{}}47.5\\ $\pm$20.8\end{tabular}                    & \begin{tabular}[c]{@{}l@{}}78.8\\ $\pm$6.9\end{tabular}          & \begin{tabular}[c]{@{}l@{}}79.8\\ $\pm$29.3\end{tabular}          & \begin{tabular}[c]{@{}l@{}}45.2\\ $\pm$10.5\end{tabular}         & \begin{tabular}[c]{@{}l@{}}74.5\\ $\pm$23.7\end{tabular}         & 68.9                                                                     \\ \hdashline
\begin{tabular}[c]{@{}l@{}}(j) Network + Verif. \\ Module + Edge Profile\end{tabular}               & \begin{tabular}[c]{@{}l@{}}80.4\\ $\pm$4.5\end{tabular}                   & \textbf{\begin{tabular}[c]{@{}l@{}}91.3\\ $\pm$3.2\end{tabular}} & \textbf{\begin{tabular}[c]{@{}l@{}}91.0\\ $\pm$2.9\end{tabular}} & \textbf{\begin{tabular}[c]{@{}l@{}}86.8\\ $\pm$7.2\end{tabular}}          & \textbf{\begin{tabular}[c]{@{}l@{}}88.4\\ $\pm$10.9\end{tabular}}          & \textbf{\begin{tabular}[c]{@{}l@{}}54.2\\ $\pm$10.0\end{tabular}}            & \textbf{\begin{tabular}[c]{@{}l@{}}75.9\\ $\pm$10.9\end{tabular}} & \textbf{\begin{tabular}[c]{@{}l@{}}68.9\\ $\pm$9.9\end{tabular}}          & \textbf{\begin{tabular}[c]{@{}l@{}}91.4\\ $\pm$4.8\end{tabular}} & \begin{tabular}[c]{@{}l@{}}48.4\\ $\pm$13.5\end{tabular}                    & \textbf{\begin{tabular}[c]{@{}l@{}}88.2\\ $\pm$3.0\end{tabular}} & \textbf{\begin{tabular}[c]{@{}l@{}}81.4\\ $\pm$28.5\end{tabular}} & \textbf{\begin{tabular}[c]{@{}l@{}}58.2\\ $\pm$4.6\end{tabular}} & \textbf{\begin{tabular}[c]{@{}l@{}}90.2\\ $\pm$9.5\end{tabular}} & \textbf{78.2}                                                                     \\ \hline
\end{tabular}
\caption{Results (mean Dice scores $\pm$ standard deviation) of different approaches on different datasets and SOIs. 
Higher value represents better performance.
In \textbf{row a}, results from both state-of-the-art methods~\cite{Isensee19, ahmad2019deep, kavur2020chaos}
and 3D UNets trained by us (values in the bracket) are reported.
Results in \textbf{row a} and \textbf{b} are only partially available in literature and they are reported just for demonstrating the approximated upper bound and limitation of fully supervised approaches,
which are not meant to be directly compared to our proposed approach.
}
\label{table:result}
\end{sidewaystable}


\section{Results and Discussion}
\label{sec:result}
The results of all the experiments are presented in Table~\ref{table:result}, 
with qualitative examples shown in 
Fig.~\ref{qualitative}
in the Supplementary Materials in Section~\ref{sec:supp}.
In Section~\ref{sec:result_auto}, 
we explore the performance change of automatic approaches in the presence of domain shift,
which leads to the analysis of the results of \proposed in Section~\ref{sec:result_semi}. 
%
%

\subsection{Automatic Approaches}
\label{sec:result_auto}
As expected, 
although the state-of-the-art performance is achieved by the \textbf{FS - Same Domain} (\textbf{row a}),
a significant performance drop (\ie over 20 Dice) can be observed 
(by comparing \textbf{row b} and the values in the brackets in \textbf{row a})
for cross-domain (\ie same SOI, different benchmarks) evaluation (\textbf{row b}).

Such variation of performance may be partially minimized by increasing the amount and diversity of training data,
better design of training augmentation,  
and application of domain adaptation techniques.
However, these may not always be practical in real-world scenarios, 
due to the high cost of data annotation and frequent domain shifts, 
for example variation of scanners and acquisition protocols in different clinical sites.

\subsection{Semi-automatic Approaches}
\label{sec:result_semi}

\proposed, 
by contrast, 
does not need any annotated data for training, but only annotation of a single slice during inference to 
indicate the SOI to be segmented.

\subsubsection{Single Slice Annotation.}
With the same amount of annotation, 
\proposed~(\textbf{row j}) clearly outperforms other baseline approaches~(\textbf{row c - f}) on all benchmarks significantly ($p<0.05$, t-test),
with an average Dice score margin of over 18.

\subsubsection{Propagation-Based Methods.}
Higher Dice score shown in \textbf{row g} over \textbf{row d - f} suggests that 
solely self-supervised correspondence matching may incur less severe error drift
and, hence, be more suitable than \textbf{Optical Flow} and \textbf{VM} for mask propagation within a volume. 
Comparison of results in \textbf{row e, f} and \textbf{g} further verifies that the backbone architecture is not the determining factor for the superior performance achieved by \proposed.
Our proposed edge profile (\textbf{row h}) is shown to be a more effective bottleneck than using the original slice as input (\textbf{row g})
and it further boosts the marginal benefit of the verification module, 
which is manifested by comparing the performance gain from \textbf{row g} to \textbf{i} and that from \textbf{row h} to \textbf{j}

\subsubsection{Self-Supervised Learning.}
Remarkably, 
\proposed trained with self-supervised learning is agnostic to SOIs and domains.
As for abdominal and chest CT, 
a \emph{single} \proposed model without any fine-tuning achieves a mean Dice score of 78.0 when testing on 7 datasets spanning 8 anatomical structures.
As for the cardiac MRI experiments with large training-testing domain shift, \proposed still performs reasonably well with a Dice score of 80.4 (\textbf{row j}).
Under this scenario, 
\proposed outperforms the fully supervised approaches significantly ($p<0.05$, t-test), 
by more than 20 Dice scores~(\textbf{row j} vs.~\textbf{row b}), 
and the annotation efforts are much lower,
\ie only a single slice per volume.

\label{sec:result}

\section{Conclusion}
In summary, 
we investigate on semi-automatic 3D segmentations, where any \emph{arbitrary} SOIs in 3D volumes are segmented by only annotating a single slice.
The proposed architecture, \proposed, is trained with self-supervised learning 
to output affinity matrices between consecutive slices through correspondence matching,
which are then used to propagate the segmentation through the volume.
Benchmarking on 8 datasets with 9 different SOIs,
\proposed shows superior generalizability and accuracy as compared to other baseline approaches, 
agnostic to the SOI. 
We envision to provide end users with more flexibility to segment and analyze different SOIs with our proposed framework,
which shows great potential to be further developed as a general interactive segmentation tool in our future works, 
to facilitate the community to study various anatomical structures,
and minimize the cost of annotating large dataset.

\label{sec:concl}

\vspace{20pt}
\par{\noindent \textbf{Acknowledgments.}} 
PH. Yeung is grateful for support from the RC Lee Centenary Scholarship. 
A. Namburete is funded by the UK Royal Academy of Engineering under its Engineering for Development Research Fellowship scheme.
W. Xie is supported by the UK Engineering and Physical Sciences Research Council (EPSRC) Programme Grant Seebibyte (EP/M013774/1) and Grant Visual AI (EP/T028572/1).  
We thank Madeleine Wyburd and Nicola Dinsdale for their valuable suggestions and comments about the work.

\bibliographystyle{splncs04}
\bibliography{ref}

\newpage
\section{Supplementary Materials}
\label{sec:supp}
\begin{figure*}[h!]
\centering
\includegraphics[width=.95\textwidth]{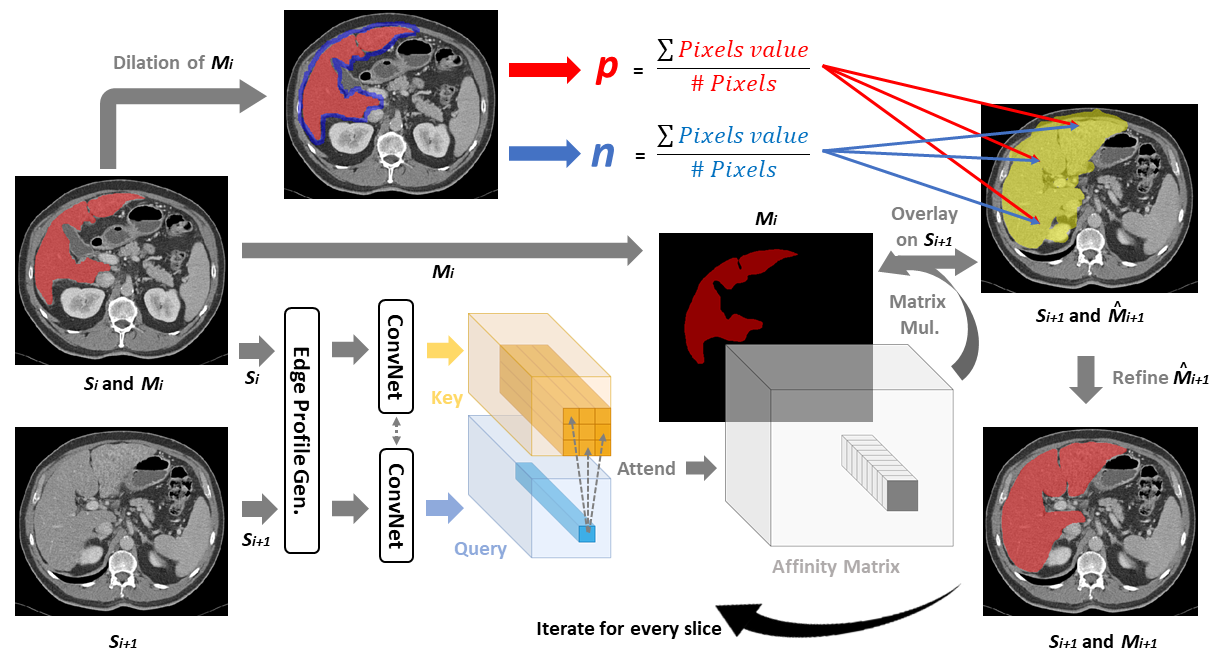}
\caption{Computation of each iteration of \proposed during \emph{inference}. 
$\{{\mathbf{S}_i}, {\mathbf{S}_{i+1}}\}$, sampled from $\mathbf{V}$ are fed into the trained correspondence flow network to obtain the affinity matrix to propagate $\mathbf{M}_i$ to $\mathbf{\hat{M}}_{i+1}$.
$\mathbf{\hat{M}}_{i+1}$ is then refined by $\mathbf{p}$ and $\mathbf{n}$, obtained by $\mathbf{M}_i$ and $\mathbf{S}_i$, to get the final mask, $\mathbf{M}_{i+1}$.} 
\label{refine}
\end{figure*}

\begin{figure*}[h!]
\centering
\includegraphics[width=.9\textwidth]{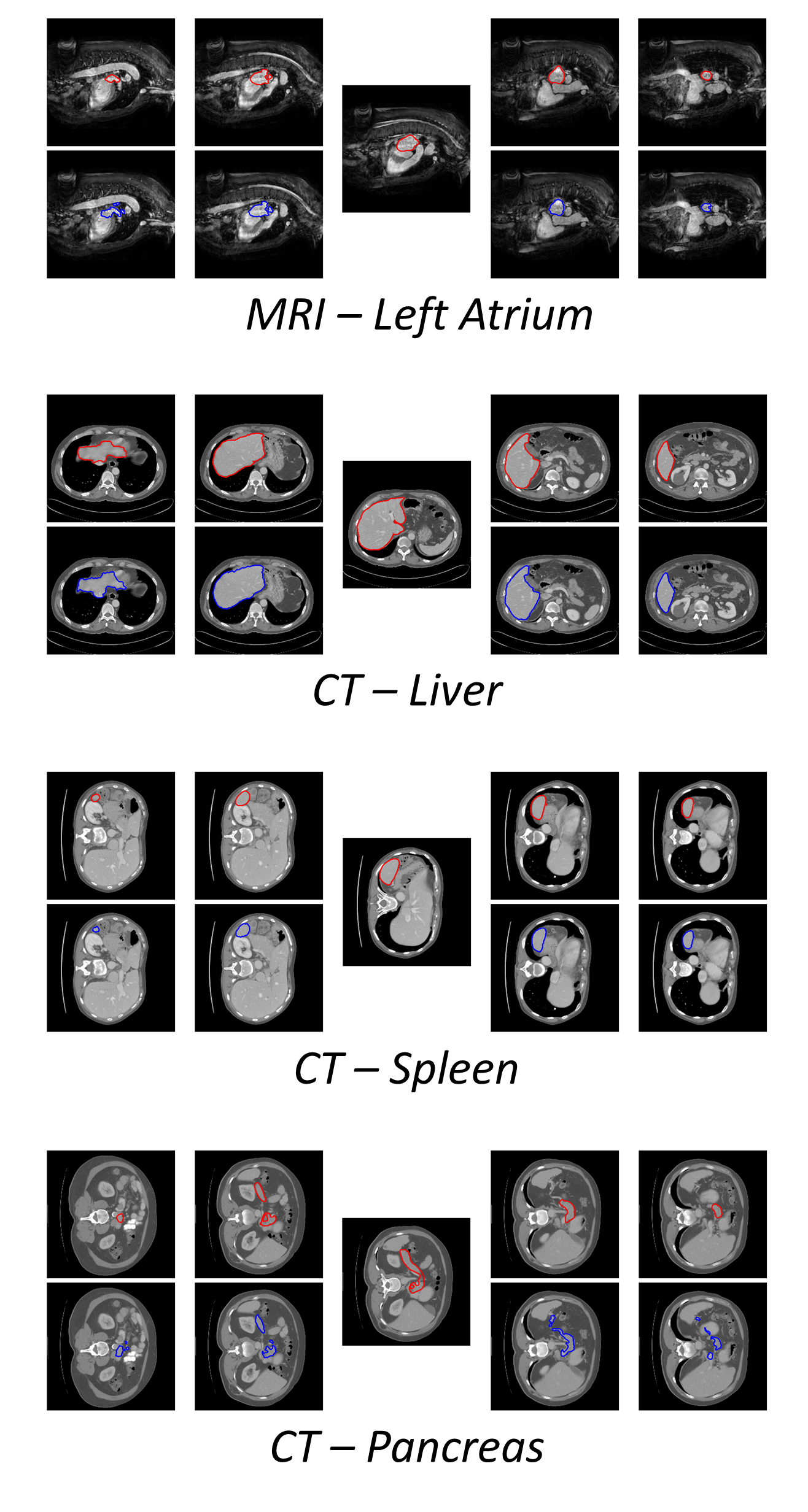}
\caption{Examples of segmentation result generated by \proposed. 
The middle slice is the initial annotation.
\textcolor{red}{Red} contours represent groundtruth segmentation while \textcolor{blue}{blue} contours represent segmentation generated by \proposed.} 
\label{qualitative}
\end{figure*}


\begin{sidewaystable}
\fontsize{7}{10}\selectfont
\begin{tabular}{|
>{\columncolor[HTML]{EFEFEF}}l |l|l|l|l|l|l|l|l|l|l|c|c|}
\hline
\textit{Modality}                                                       & \multicolumn{10}{c|}{Abdominal and Chest CT}                                                                                                                                                                                                                                                                                                                                                                                                                                                                                                                                                                               & \multicolumn{2}{c|}{Cardiac MRI}                                                             \\ \hline
\textit{Task}                                                           & \multicolumn{3}{c|}{\textbf{Training}}                                                                                                                                               & \multicolumn{7}{c|}{\textbf{Testing}}                                                                                                                                                                                                                                                                                                                                                                                               & \textbf{Training}       & \textbf{Testing}                                                   \\ \hline
\textit{Name}                                                           & \multicolumn{1}{c|}{\begin{tabular}[c]{@{}c@{}}C4KC-\\ KiTS\end{tabular}} & \multicolumn{1}{c|}{CT-LN} & \multicolumn{1}{c|}{\begin{tabular}[c]{@{}c@{}}CT-\\ Pancreas\end{tabular}} & \multicolumn{1}{c|}{Sliver07} & \multicolumn{1}{c|}{CHAOS}                                                                      & \begin{tabular}[c]{@{}l@{}}Decath-\\ Liver\end{tabular} & \begin{tabular}[c]{@{}c@{}}Decath-\\ Spleen\end{tabular} & \begin{tabular}[c]{@{}l@{}}Decath-\\ Pancreas\end{tabular} & \begin{tabular}[c]{@{}c@{}}3Dircadb\\ -01\end{tabular} & \begin{tabular}[c]{@{}c@{}}3Dircadb\\ -02\end{tabular} & Kaggle                  & \begin{tabular}[c]{@{}l@{}}Decath\\ -Heart\end{tabular}            \\ \hline
\textit{Type}                                                           & \multicolumn{10}{l|}{3D Volumes}                                                                                                                                                                                                                                                                                                                                                                                                                                                                                                                                                                                           & 2D Video Sequence       & 3D Volume                                                          \\ \hline
\textit{SOI}                                                            & \multicolumn{1}{c|}{-}                                                    & \multicolumn{1}{c|}{-}     & \multicolumn{1}{c|}{-}                                                      & Liver                         & Liver                                                                                           & Spleen                                                  & Liver                                                    & Pancreas                                                   & Multiple                                               & Multiple                                               & \multicolumn{1}{c|}{-}  & Left atrium                                                        \\ \hline
\textit{Number}                                                         & 310                                                                       & 86                         & 82                                                                          & 20                            & 20                                                                                              & 41                                                      & 131                                                      & 281                                                        & 20                                                     & 2                                                      & 14370                   & 20                                                                 \\ \hline
\textit{Scanner}                                                        & Multiple                                                                  & Multiple                   & \begin{tabular}[c]{@{}l@{}}Philips \& \\ Siemens\\ MDCT\end{tabular}        & Multiple                      & \begin{tabular}[c]{@{}l@{}}Philips Secura \\ Philips Mx8000 \\ Toshiba AquilionOne\end{tabular} & \multicolumn{1}{c|}{NA}                                 & Multiple                                                 & \multicolumn{1}{c|}{NA}                                    & \multicolumn{1}{c|}{NA}                                & \multicolumn{1}{c|}{NA}                                & \multicolumn{1}{c|}{NA} & \begin{tabular}[c]{@{}l@{}}Philips \\ 1.5T \\ Achieva\end{tabular} \\ \hline
\textit{\begin{tabular}[c]{@{}l@{}}Resolution\\ (xy) (mm)\end{tabular}} & Varying                                                                   & Varying                    & Varying                                                                     & 0.55-0.8                      & 0.7-0.8                                                                                         & Varying                                                 & 0.5-1.0                                                  & Varying                                                    & Varying                                                & Varying                                                & Varying                 & 1.25                                                               \\ \hline
\textit{\begin{tabular}[c]{@{}l@{}}Resolution\\ (z) (mm)\end{tabular}}  & Varying                                                                   & Varying                    & 1.5-2.5                                                                     & 1.0-3.0                       & 3.0-3.2                                                                                         & 2.5-5.0                                                 & 0.45-6.0                                                 & 2.5                                                        & Varying                                                & Varying                                                & \multicolumn{1}{c|}{-}  & 2.7                                                                \\ \hline
\textit{Details}                                                        &                                   \cite{c4kc}                                        &              \cite{lymphnodes}              &                                                                            \cite{pancreasct} &               \cite{van07}                &                                          \cite{CHAOSdata2019}                                                       &\cite{simpson2019large}                                                        &        \cite{simpson2019large}                                                  &\cite{simpson2019large}                                                            &                     \cite{soler20103d}                                   &                   \cite{soler20103d}                                     &            \cite{kaggle}             &\cite{simpson2019large}                                                                    \\ \hline
\end{tabular}
\caption{Summarization of different datasets used for our experiments. 
}
\label{table:data}

\vspace*{4mm}
\begin{tabular}{|
>{\columncolor[HTML]{EFEFEF}}l |c|l|l|l|l|l|l|l|}
\hline
\textit{Approaches}                                                                                                       & \multicolumn{2}{l|}{\textbf{\begin{tabular}[c]{@{}l@{}}Fully Supervised - \\ Same Domain\end{tabular}}}                                                               & \textbf{\begin{tabular}[c]{@{}l@{}}Fully Supervised - \\ Different Domain\end{tabular}} & \textbf{\begin{tabular}[c]{@{}l@{}}Fully Supervised - \\ Single Slice\end{tabular}} & \textbf{Optical Flow}                                                                                                & \textbf{\begin{tabular}[c]{@{}l@{}}VoxelMorph2D - \\ UNet\end{tabular}}            & \textbf{\begin{tabular}[c]{@{}l@{}}VoxelMorph2D -\\ ResNet18Stride1\end{tabular}} & \proposed                                                                               \\ \hline
\textit{\begin{tabular}[c]{@{}l@{}}Backbone\\ Architecture\end{tabular}}                                                  & Varying                                                                         & \multicolumn{2}{l|}{\begin{tabular}[c]{@{}l@{}}- 3D Unet\\ - 16 filters at first level\end{tabular}}                                                                          & \begin{tabular}[c]{@{}l@{}}- 2D Unet\\ - 64 filters \\ at first level\end{tabular}  & \begin{tabular}[c]{@{}l@{}}conventional off-\\ the-shelf optical \\ flow algorithm~\cite{farneback2003two}\end{tabular}                      & \begin{tabular}[c]{@{}l@{}}- 2D Unet\\ - 64 filters \\ at first level\end{tabular} & \multicolumn{2}{l|}{\begin{tabular}[c]{@{}l@{}}- ResNet18 without max pooling \\ and stride at every layer equals 1\\ - 16 filters at first level\end{tabular}}                    \\ \hline
\cellcolor[HTML]{EFEFEF}                                                                                                  &                                                                                 & \multicolumn{2}{l|}{- Batch size of 1}                                                                                                                                        & - Batch size of 10                                                                  & \multicolumn{1}{c|}{}                                                                                                & \multicolumn{3}{l|}{- Batch size of 10}                                                                                                                                                                                                                                 \\ \cline{3-5}
\multirow{-2}{*}{\cellcolor[HTML]{EFEFEF}\textit{\begin{tabular}[c]{@{}l@{}}Training \\ hyper-\\ parameter\end{tabular}}} & \multirow{-2}{*}{Varying}                                                       & \multicolumn{3}{l|}{\begin{tabular}[c]{@{}l@{}}- Learning rate (lr) of 0.0001\\ - lr halved when errors plateaued\\ - ADAM optimization\end{tabular}}                                                                                                               & \multicolumn{1}{c|}{\multirow{-2}{*}{-}}                                                                             & \multicolumn{3}{l|}{\begin{tabular}[c]{@{}l@{}}- Learning rate (lr) of 0.0001\\ - lr halved every epoch\\ - ADAM optimization\end{tabular}}                                                                                                                             \\ \hline
\textit{\begin{tabular}[c]{@{}l@{}}Input dim.\end{tabular}}                                                        & Varying                                                                         & \multicolumn{2}{c|}{(128, 128, 128)}                                                                                                                                          & \multicolumn{5}{c|}{(256, 256)}                                                                                                                                                                                                                                                                                                                                                                                                                                                      \\ \hline
\textit{Remarks}                                                                                                          & \multicolumn{1}{l|}{\begin{tabular}[c]{@{}l@{}}Results from\\ \cite{Isensee19, ahmad2019deep, kavur2020chaos, tran2020multiple}\end{tabular}} & \begin{tabular}[c]{@{}l@{}}Results from\\ model trained\\ by ourselves\end{tabular} & \multicolumn{1}{c|}{-}                                                                  & \multicolumn{1}{c|}{-}                                                              & \begin{tabular}[c]{@{}l@{}}Hyperparameters \\ from OpenCV~\cite{opencv_library}:\\ pyr\_scale = 0.5\\ level = 3\\ winsize = 7\end{tabular} & \multicolumn{1}{c|}{-}                                                             & \multicolumn{1}{c|}{-}                                                            & \begin{tabular}[c]{@{}l@{}}Other hyper-\\ parameters:\\ d = 8\\ s = 3\\ w = 15x15\end{tabular} \\ \hline
\end{tabular}
\caption{Implementation details of \proposed and other basline approaches.
}
\label{table:implementation}
\end{sidewaystable}


\label{sec:supp}

\end{document}